\documentclass[letterpaper, 10 pt, conference]{ieeeconf2}  

\IEEEoverridecommandlockouts                              

\usepackage{cite}
\usepackage{amsmath,amssymb,amsfonts}
\usepackage{algorithmic}
\usepackage{graphicx}
\usepackage{textcomp}
\usepackage{xcolor}
\usepackage{multirow}
\usepackage{color}
\usepackage{arydshln}

\title{\LARGE \bf
Towards Generalizable Referring Image Segmentation via \\Target Prompt and Visual Coherence\\ 
}

\author{ 
Yajie Liu$^{1}$, Pu Ge$^{2}$,  Haoxiang Ma$^{1}$, Shichao Fan$^{3}$, Qingjie Liu$^{1,2}$, Di Huang$^{1,2}$, Yunhong Wang$^{1,2}$  
\thanks{$^{1}$ School of Computer Science and Engineering, Beihang University, Beijing, China.}
\thanks{$^{2}$ Hangzhou Innovation Institute, Beihang University, Beijing, China}%
\thanks{$^{3}$ School of Mechanical Engineering and Automation, Beihang University, Beijing, China}
}

\begin{document}

\maketitle
\thispagestyle{empty}
\pagestyle{empty}

\begin{abstract}

Referring image segmentation (RIS) aims to segment objects in an image conditioning on free-from text descriptions. Despite the overwhelming progress, it still remains challenging for current approaches to perform well on cases with various text expressions or with unseen visual entities, limiting its further application. In this paper, we present a novel RIS approach, which substantially improves the generalization ability by addressing the two dilemmas mentioned above. Specially, to deal with unconstrained texts, we propose to boost a given expression with an explicit and crucial prompt, which complements the expression in a unified context, facilitating target capturing in the presence of linguistic style changes. Furthermore, we introduce a multi-modal fusion aggregation module with visual guidance from a powerful pretrained model to leverage spatial relations and pixel coherences to handle the incomplete target masks and false positive irregular clumps which often appear on unseen visual entities. Extensive experiments are conducted in the zero-shot cross-dataset settings and the proposed approach achieves consistent gains compared to the state-of-the-art, e.g., 4.15\%, 5.45\%, and 4.64\% mIoU increase on RefCOCO, RefCOCO+ and ReferIt respectively, demonstrating its effectiveness. Additionally, the results on GraspNet-RIS show that our approach also generalizes well to new scenarios with large domain shifts.

\end{abstract}

\section{INTRODUCTION}

Referring image segmentation (RIS) aims to segment objects in an image described by a language expression. It has a broad range of applications such as language-based human-robot interaction~\cite{wang2019reinforced}, image editing~\cite{chen2018language}, etc. In contrast to the conventional semantic/instance segmentation~\cite{zhou2019semantic} that segments objectives in a fixed set of categories, the targets of RIS are given by free-from expressions with much richer vocabularies and syntactic varieties. An algorithm for RIS needs to infer the target from the given expression and segment the corresponding visual entities in the image. Existing state-of-the-art methods have achieved remarkable performance on multiple benchmarks. However, there remain many challenges that need to be solved to achieve the intention of this task. 

 \begin{figure}[h]
\centerline{\includegraphics[width=0.45\textwidth]{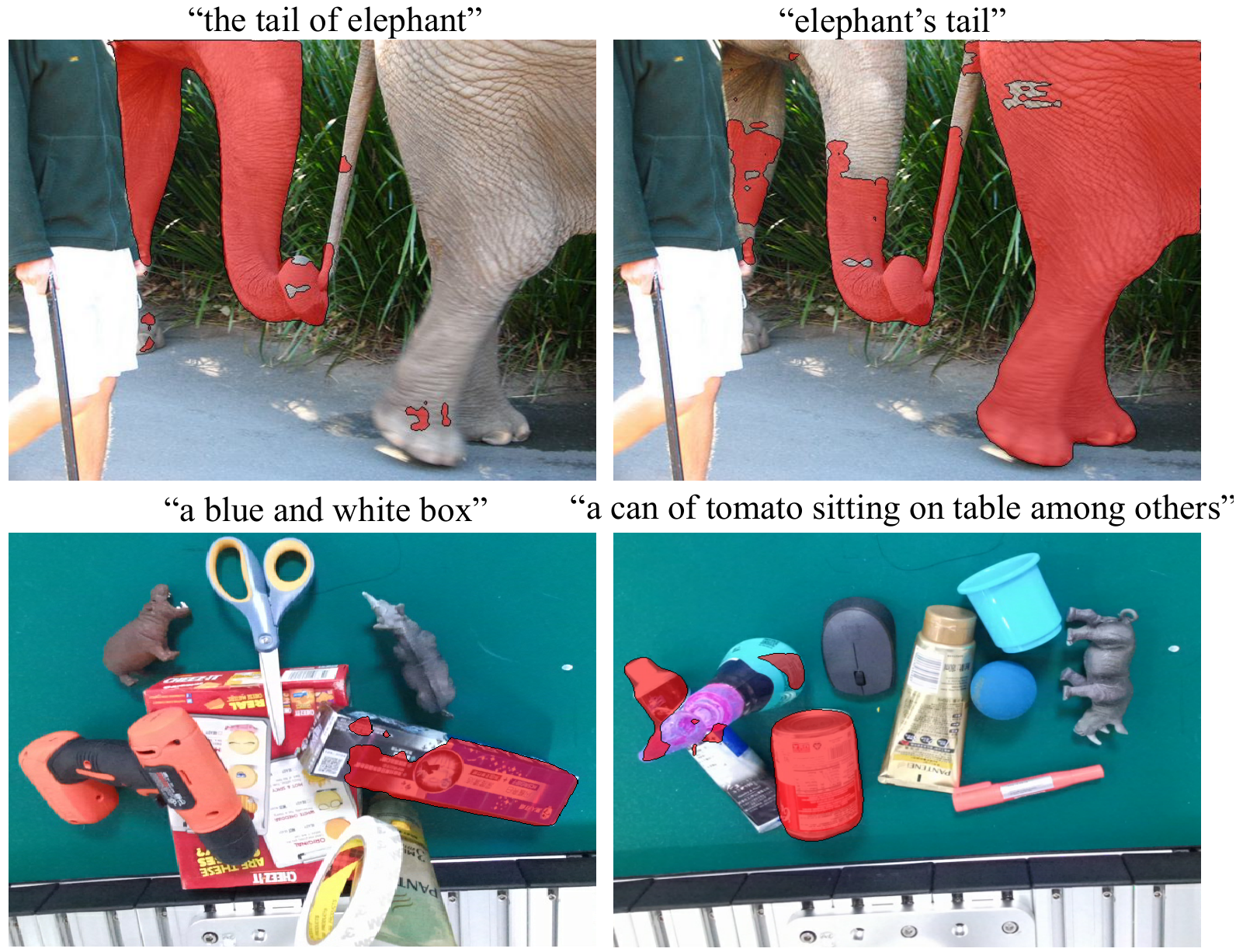}}
\caption{Predictions from the model pretrained on RefCOCO released by LAVT. The first row shows the predicted masks for the same object described by different text descriptions; The second row shows the results for GraspNet, examples of incomplete target masks and false positive clumps.}

\vspace{-0.4cm} 
\label{fig1}
\end{figure}

Firstly,~\textbf{\emph{can a well-trained model robustly generalize to unseen language expressions with varied syntaxes or descriptions of different characteristics of the object}}?  
Regrettably, the answer is no.
As an example demonstrated in Fig.~\ref{fig1}, we describe the elephant tail by two different expressions, then feed them into the model pretrained on RefCOCO~\cite{yu2016modeling} (72.73 oIoU on the validation set) released by LAVT~\cite{yang2022lavt}, which we denote as LMODEL.  Surprisingly, the two predicted masks are completely different as shown in the first row of Fig.~\ref{fig1}. 
Moreover, the performance of LMODEL on RefCOCO+~\cite{yu2016modeling} decreases significantly, i.e., 72.73 oIoU on RefCOCO \emph{v.s.} 56.86 oIoU on RefCOCO+, even though they share same image domain and object categories.

Another challenge is,~\textbf{\emph{can a well-trained model effectively generalize to unseen visual entities including images with domain shift and objects of unseen categories}}?  
We observe that the LMODEL exhibits poor performance on ReferIt~\cite{kazemzade2014referring} (21.22 oIoU) which comprises quite different images and categories compared to RefCOCO.
 Moreover, we test the model on GraspNet~\cite{fang2020graspnet}, a general object grasping dataset in robotics, as shown in the second row of Fig.~\ref{fig1}. There are two typical defects in the predictions: incomplete target masks and false positive clumps, hindering the generalization of RIS in practical applications. 
As discussed above, existing state-of-the-art RIS methods struggle to ``understand'' the various free-form text expressions and perform poorly on unseen visual entities, limiting further application. Collecting data for each domain is extremely expensive which demands exhaustive annotations of precise referring expressions and pixel-wise masks.  Moreover, it is impossible to cover all potential textual expressions for visual entities, even within a fixed set of categories.

In this paper, we aim to enhance the generalization of RIS by addressing the two challenges mentioned above.
Firstly, in order to generalize well to the unconstrained textual descriptions, it is essential to robustly capture the true target without biasing towards disturbing items in the expression. To achieve this, we propose to boost a given expression with an explicit and crucial prompt. The prompt complements the expression in a unified context with the core of the target noun extracted from the expression. Based on the unified form, the model can effectively capture the target, which can reduce spurious visual-text correlations and promote knowledge learning across instances with identical prompts. Additionally, the model can learn characteristics relevant to the target in the descriptions as ``residuals'' with the explicit target prompt.

The reason behind the second challenge lies in the pixel-word alignment in existing methods. 
They connect textual features with visual features at each pixel respectively to learn effective cross-modal alignments. However, this inevitably leads to inconsistent predictions within an object, especially in unseen visual entities as shown in Fig.~\ref{fig1}. To handle this issue, we introduce a Multi-modal Fusion Aggregation (MFA) module to leverage the visual coherences based on the well-established generalization capabilities of aggregation modules in matching literature~\cite{liu2022graftnet}.
Furthermore, we incorporate the visual knowledge from a powerful pretrained vision model inspired by ~\cite{cho2023cat} to guide the MFA and the decoding of cross-modal features.

To validate the effectiveness of our method, we propose a zero-shot cross-dataset evaluation protocol similar to the open-vocabulary settings~\cite{xu2022simple}.
Comprehensive results demonstrate that our method consistently enhances the generalization of RIS across all settings. And we produce the GraspNet-RIS by generating text descriptions for GraspNet using the multi-modal LLM  Shikra~\cite{chen2023shikra}. The test results on it show our approach also generalizes well to practical applications.

Our main contributions can be summarised as follows:
\begin{itemize}
\item We introduce a novel task of zero-shot cross-dataset RIS to assess the generalization of the model to various unconstrained expressions and unseen visual entities, including images with domain shift and unseen categories.
\item We boost the model's generalization to various text descriptions by supplementing the given expression with an explicit target prompt in a unified context. We introduce a multi-modal fusion aggregation module along with visual knowledge guidance to deal with the incomplete target masks and false positive clumps that often appear on unseen visual entities.
\item We conduct comprehensive experiments and our method achieves consistent gains compared to the state-of-the-art RIS methods, demonstrating its effectiveness.
\end{itemize}

\section{RELATED WORK}

\subsection{Referring image segmentation}

Referring image segmentation has attracted growing attention in the research community with numerous studies yielding significant results~\cite{feng2021encoder},~\cite{chen2022position},~\cite{sima2023embodied},~\cite{chen2021referring}.
Most research works focus on getting effective cross-modal fusion representations. These efforts have evolved from basic fusion approaches, such as concatenation~\cite{kamath2021mdetr}, to more advanced techniques, such as the utilization of transformer blocks~\cite{ding2021vision}, from decoder-only late fusion to early fusion in the encoder~\cite{yang2022lavt}.
Especially,~\cite{yang2022lavt} leverages a hierarchical vision Transformer to jointly embed language and vision features to facilitate cross-modal alignments.~\cite{zhang2022coupalign} combines sentence-mask alignment with word-pixel alignment to hierarchically align pixels and masks with linguistic cues.
However, the test results above indicate their generalization abilities do not meet expectations.

Recently, several studies~\cite{liu2023polyformer},~\cite{zhu2022seqtr} utilize a Transformer-based encoder-decoder architecture~\cite{vaswani2017attention} and transform RIS into the predictions of target point sets. Nevertheless, achieving competitive results with this architecture demands a substantial amount of pre-training data due to its complexity.
In contrast to these works, we aim to enhance the generalization of RIS without using additional training data.

\subsection{Generalization of Referring Image Segmentation}
To the best of our knowledge, few works have studied the generalization of RIS.~\cite{subramanian2022reclip} observes that a model trained on one dataset exhibits poor performance when applied to a different image domain. They focus on zero-shot referring expression comprehension using large-scale pretrained vision and language models such as CLIP~\cite{radford2021learning}, ALIGN~\cite{jia2021scaling} to avoid the domain shift in supervised learning. 
Similarly,~\cite{yu2023zero} transfers from the pretrained knowledge of CLIP to zero-shot RIS. They exploit an unsupervised mask generator to generate mask proposals and recognize the target mask by computing the similarity map using CLIP.
Nonetheless, a substantial gap remains between zero-shot performance and supervised RIS, 24.88 oIoU on the RefCOCO validation set in~\cite{yu2023zero}~\emph{v.s.} 72.73 oIoU in~\cite{yang2022lavt}. 

CGformer \cite{tang2023contrastive} evaluates the generalization of RIS in zero-shot cross-class settings across the RefCOCO series datasets. In contrast, we propose a more challenging task, where the model not only deals with unseen classes but also has to tackle various text expressions and address the domain gap among different datasets.

\begin{figure*}[htbp]
\setlength{\abovecaptionskip}{0pt}
\setlength{\belowcaptionskip}{-0pt}
\vspace{5pt}
\begin{center}
\centerline{\includegraphics[width=0.9\textwidth]{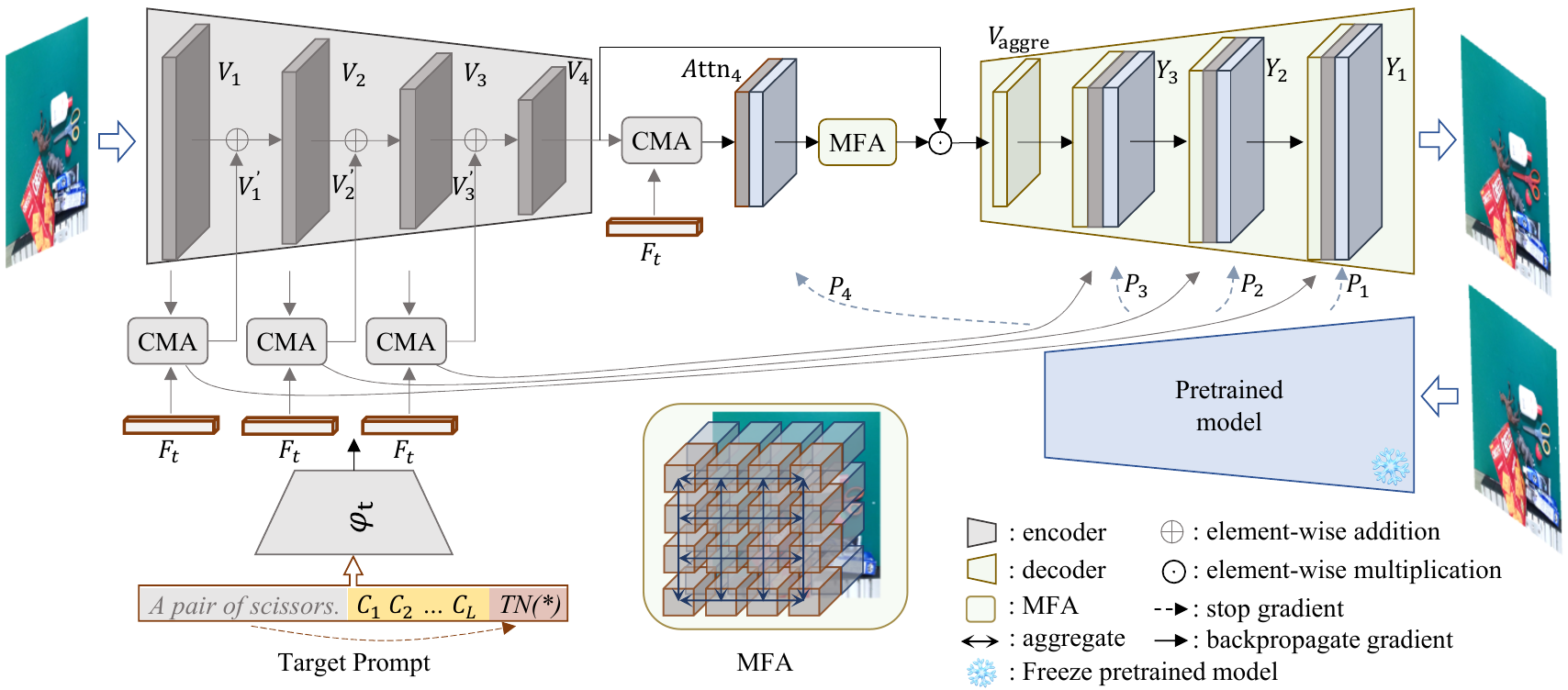}}
\caption{The overall pipeline of our proposed method. Given a text description, we complement the expression with target prompt in a unified context, then feed it into text encoder. The visual encoding process is the same as LAVT. We introduce MFA and visual knowledge from a pretrained model to guide the multi-model fusion and decoding process. The gray parts are identical to those in LAVT. CMA: cross-modal alignment. MFA: multi-modal fusion aggregation.}
\label{fig3}
\end{center}
\vspace{-0.5cm} 
\end{figure*}
\section{PROPOSED APPROACH}
In this paper, we propose to enhance the generalization of RIS from two perspectives.
Firstly, we complement the given expression with an explicit target prompt in a unified form to address the challenge of various free-from text descriptions.
Secondly, a multi-modal fusion aggregation module along with visual guidance from a powerful pretrained model is introduced. It exploits visual coherences to tackle incomplete target masks and noisy clumps that often appear on unseen visual entities which result in poor performance. 
We apply the aforementioned improvements to LAVT~\cite{yang2022lavt}, constructing a strong baseline for the generalization of RIS. 
The details of the overall method are illustrated in Fig.~\ref{fig3}.
\subsection{Problem Formulation}
We briefly review LAVT, the first model that conducts cross-modal feature fusion in the early encoding stage, achieving excellent results in multiple benchmarks. 
Given an image $I$ and the referring expression $ E_{r}$, LAVT forwards $ E_{r}$ to the text encoder BERT~\cite{kenton2019bert} to get the textual feature  $F_{t}\in\mathbb{R}^{C_{t} \times T}$, where $C_{t}$ and $T$ denote the number of channels for text features and number of tokens, respectively. 
To align the given expression to visual pixels, LAVT jointly embeds the image features $V_{i} \in \mathbb{R}^{C_{i} \times H_{i} \times W_{i}} $, $i \in \{1,2,3,4\} $ from the visual encoder Swin-Transformer~\cite{liu2021swin} with the guidance of pixel-word attention module (PWAM).
Based on the cross-modal features, a lightweight mask decoder is employed to obtain the final mask. We propose an improved LAVT that robustly generalizes to unseen referring expressions and visual entities.

\subsection{Target Prompt}

The text expression provides a distinguishing description of the targets in an image that needs to be segmented. 
As demonstrated in the realistic human-robot interaction collaboration experiment presented in~\cite{shridhar2018interactive}, referring expressions are unconstrained and vary significantly among human users. Even a simple object, such as a tomato can, is described in many different ways as ``the can in the middle'', ``a tomato can'', ``the second can from the right on the top row'', etc. 
Without providing direct and explicit guidance, identifying the target robustly in various referring expressions is an impossible task for a text encoder which is primarily supervised to extract textual representations aligned with visual features with limited training data.

To address the above issues, we propose to guide the text encoder with an explicit and crucial prompt. We first exploit a dependency parsing tool~\cite{honnibal2015improved} to parse the given expression $ E_{r}$ into a tree structure and identify the root noun of the sentence as the target $TN(E_{r})$.
After capturing the target in the referring expression, we adopt it to boost the text expression in a unified context to get the final referring expression~\emph{$E_{f}$}. The unified context for the prompt is designed with the following form inspired by~\cite{zhou2022learning}:
\begin{equation}
E_{f}=[E_{r}][C]_{1}[C]_{2}...[C]_{L}[TN(E_{r})]\label{prompt}
\end{equation}
where each $[C]_{l}~(l \in{1, ..., L}$) is a token in the unified context. $L$ is the number of context tokens. 

\noindent\textbf{Manual Prompt} We customize the prompt context for RIS with the form of ``It is a [$TN(E_{r})$]'', which is shared among all referring expressions. Given an expression~\emph{$ E_{r}$}, we obtain the referring expression~\emph{$E_{f}$} via: 
\begin{equation}
E_{f}=[E_{r}] .\ It\ is\ a\ [TN(E_{r})]\label{manu_prompt}
\end{equation}

It should be noted that the unified context can be formed in various ways, e.g., ``It describes a [$TN(E_{r})$]'',``It is one of a [$TN(E_{r})$]'', etc. The learnable prompt is not employed here because the context embeddings need to be learned from the training data, which may hinder the generalization of the method.

Based on the unified context, the model can effectively capture the target within the expression regardless of its linguistic style, thereby reducing spurious visual-text correlations caused by biasing towards disturbing items in the expression. 
Moreover, it can facilitate knowledge learning across instances with identical prompts, i.e., the same $TN(E_{r})$.
By forwarding~\emph{$E_{f}$} to text encoder~$\varphi_{t}$, we obtain the textual feature $F_{t}$:
\begin{equation}
F_{t}=\varphi_{t}(E_{f})\label{extract_tfea}
\end{equation}

\subsection{Multi-modal Fusion Aggregation and Visual Guiding}
In order to align the visual pixels with text features to distinguish referred object from the entire image, pixel-word attention is employed to enhance pixel-word alignment in recent RIS methods~\cite{yang2022lavt},~\cite{zhang2022coupalign}. Given the visual feature  $V_{i} \in \mathbb{R}^{C_{i} \times H_{i} \times W_{i}} $ and textual feature $F_{t} \in \mathbb{R}^{C_{t} \times T } $, a cross-attention layer is utilized to fuse multi-modal features computed as:
\begin{equation}
Attn_{i}=softmax(W_{iv}(V_{i})W_{t}(F_{t}))\label{attn}
\end{equation}
\begin{equation}
{V_{i}}^{'} = Attn_{i} \odot W_{im}(V_{i})\label{vi}
\end{equation}
where $\odot$ denotes element-wise multiplication and $W_{iv}$, $W_{t}$, $W_{im}$ are projection layers. ${V_{i}}^{'}$ is the visual feature modulated by the pixel-word attention, i.e., the multi-modal fusion feature, which is the input of the mask decoder.

\begin{figure}[h]
\setlength{\abovecaptionskip}{0pt}
\setlength{\belowcaptionskip}{-0pt}
\vspace{5pt}
\centerline{\includegraphics[width=0.48\textwidth]{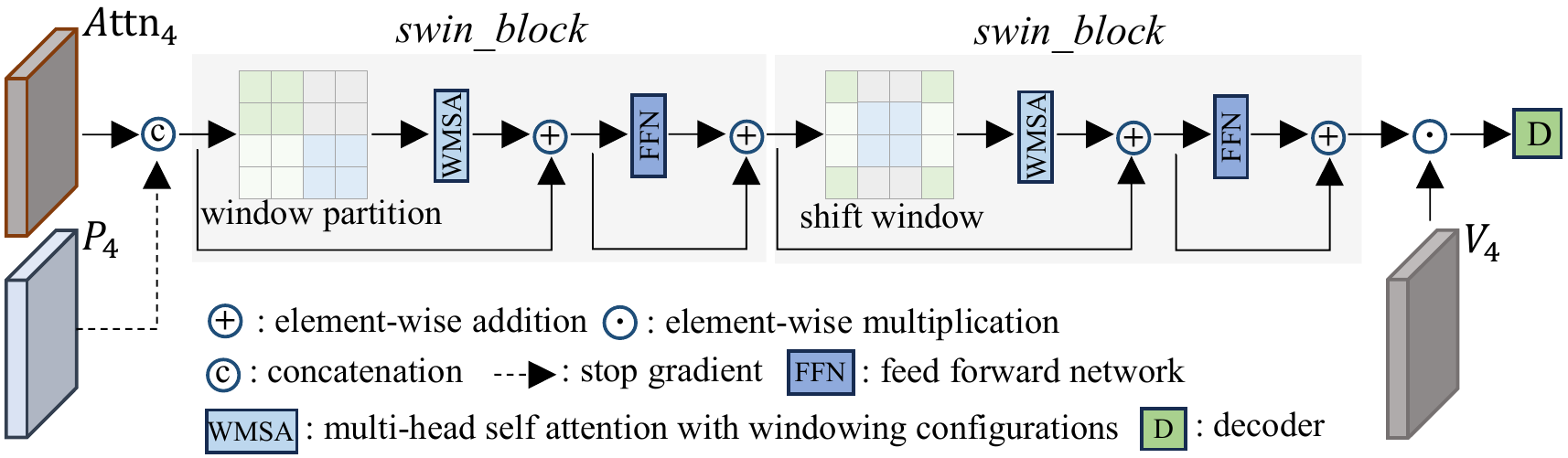}}
\caption{Pipeline of the multi-modal fusion aggregation module (MFA). We omit LayerNorm in swin\_blocks for a clearer view.}
\label{mfa}
\vspace{-0.4cm} 
\end{figure}

The defect of pixel-word alignment is that it considers the similarity between each pixel embedding and text feature independently, ignoring the semantic consistency within an object. As a result, there would be irregular false positive clumps or missing part of target pixels on the predictions, especially in unseen data without supervision, leading to poor generalization. To solve the above issues, we introduce a multi-modal fusion aggregation (MFA) module into cross-attention layer to capture the spatial relations and harness the favorable generalization capabilities of aggregation module. Concretely, we feed the pixel-word attention $Attn$ into the MFA module, as shown in Fig.~\ref{mfa}. Notably, we only introduce MFA into the final cross-attention module whose fused features are fed into the decoder for computational efficiency.

The aggregation module needs to have semi-global receptive field to capture the spatial relations among local regions. Similar to the cost aggregation in~\cite{cho2023cat}, the introduced MFA contains two consecutive Swin-Transformer blocks. Each block models the spatial relations within a local window via self-attention. 
As the text features are same for each pixel, visual similar regions have similar multi-modal fusion features.  Based on this insight, we incorporate visual guidance from a powerful pretrained vision model to guide the aggregation of multi-modal features with the visual spatial relations.
Notably, different from~\cite{cho2023cat} which leverages the aggregation in the final cost of visual features and textual features, we introduce the aggregation into the multi-modal fusion during encoding. Similarly, we also guide the decoding process of multi-modal fusion features with visual knowledge. 

Given the pretrained vision model, we extract the visual embeddings $P_i$, $i\in \{1, 2, 3, 4\}$. MFA can be computed as:
\begin{equation}
V_{aggre} = swin\_blocks(Attn_{i};W_i(P_i))\odot W_{im}(V_{i})\label{aggre}
\end{equation}
where $W_i$ denotes a linear projection layer, [;] denotes concatenation along the channel dimension. The decoding process can be described as:
\begin{equation}
\begin{cases}
\, Y_4 = V_{aggre}\\
\, Y_i = \rho_i([Up(Y_i+1); V_i^{'}; W_p(sg(P_i))]), i=3,2,1
\label{decoder}
\end{cases}
\end{equation}

where $\rho_i$ is a projection function similar to~\cite{yang2022lavt}.
We apply group normalization~\cite{wu2018group} instead of batch normalization~\cite{ioffe2015batch} in $\rho_i$ to better capture channel features from multiple sources. $Up$ represents upsampling via bilinear interpolation.
$W_p$ denotes a linear projection layer. $sg$ is the stop gradient operator. The final feature map $Y_1$ is fed into the classification head (foreground and background) to obtain the prediction mask.

\section{EXPERIMENTS}
\subsection{Datasets and Evaluation Protocol}
\noindent\textbf{Datasets.} There are four widely-used datasets for referring image segmentation: RefCOCO \cite{yu2016modeling}, RefCOCO+~\cite{yu2016modeling}, RefCOCOg~\cite{mao2016generation},\cite{nagaraja2016modeling} and ReferIt \cite{kazemzade2014referring}. The RefCOCO series datasets are collected on top of Microsoft COCO~\cite{lin2014microsoft}, which comprises 80 common categories.
The referring expressions for RefCOCO and RefCOCO+ are both collected using the ReferitGame \cite{kazemzade2014referring}. The difference is that RefCOCO+ focuses solely on appearance and is disallowed from using location words, while no restrictions are placed on  RefCOCO expressions. RefCOCOg expressions are collected on Amazon Mechanical Turk 
and the descriptions are longer and more complex (8.4 words on average~\emph{vs}. 3.61 words of RefCOCO and 3.53 words of RefCOCO+). ReferIt is collected from the ImageCLEF IAPR dataset~\cite{grubinger2006iapr},
which contains 238 different categories. Obviously, there exists a significant domain gap in terms of images between ReferIt and the other three datasets. Furthermore, it contains a considerable number of categories that do not appear in the RefCOCO series datasets.

To evaluate the generalization of RIS in the context of human-robot interaction, we generate referring expressions for a subset of images from GraspNet~\cite{fang2020graspnet} using Shikra~\cite{chen2023shikra}. We collect the first frame for every scene from the subsets train\_1, test\_seen, test\_similar, and test\_novel. Subsequently, we select objects with clear class definitions and typical appearance, resulting in 50 categories. For instance, we exclude the~\emph{072-k\_toy\_airplane} which looks completely different from a conventional airplane. We generate the text expressions as follows:
\begin{itemize}
    \item Select objects belong to chosen categories and calculate their bounding boxes based on the instance masks.
    \item Feed the image and each bounding box to Shikra with three different prompts, resulting in three referring expressions for each object.
    \item Pick the most precise and diverse description and make adjustments to ensure it accurately corresponds to the object if necessary.
\end{itemize}

In total, we generate 497 referring expressions for 105 images from GraspNet, which we denote as GraspNet-RIS. 

\begin{table*}[ht]
\setlength{\abovecaptionskip}{0pt}
\setlength{\belowcaptionskip}{-0pt}
\vspace{5pt}
\caption{Zero-shot cross-dataset performance comparison with state-of-the-art methods on public datasets. The models are all trained on one dataset and evaluated on the other datasets.  The test results of the trained dataset are~\textbf {in gray}.}
\centering
\begin{tabular}{c|c|c|c c c|c c c|c c|c c}
\hline
\multirow{2}*{Train Dataset}
& \multicolumn{2}{c|}{\multirow{2}*{Method}}
&\multicolumn{3}{c|}{RefCOCO}
&\multicolumn{3}{c|}{RefCOCO+}
&\multicolumn{2}{c|}{RefCOCOg}
&\multicolumn{2}{c}{ReferIt} \\
\cline{4-13} 
 &\multicolumn{2}{c|}{} & val & testA & testB  & val & testA & testB  & val & test & val & test \\
\hline

\multirow{6}*{RefCOCO}
& \multirow{3}*{oIoU}
& VLT~\cite{ding2021vision}
& \textcolor{gray}{63.15} & \textcolor{gray}{66.57} & \textcolor{gray}{59.75}
& 45.65 & 50.21 & 39.14
& 49.81 & 49.52
& 18.56 & 28.52 \\
& & LAVT~\cite{yang2022lavt}
&  \textcolor{gray}{72.73} & \textcolor{gray}{ 75.82} & \textcolor{gray}{68.79}
 & 56.86 & 62.29  & 48.14
&  59.76  & 58.65
 & 21.22  & 35.35 \\
& & Ours & \textcolor{gray}{73.43}  &\textcolor{gray}{76.40}  &\textcolor{gray}{72.05} 
    &\textbf{57.68}  & \textbf{63.71}  & \textbf{49.76} 
   & \textbf{60.32}  & \textbf{61.70}
   & \textbf{22.97}  & \textbf{37.13} \\
\cline{2-13}
& \multirow{3}*{mIoU}
& VLT~\cite{ding2021vision}
&\textcolor{gray}{66.15} &\textcolor{gray}{68.95} &\textcolor{gray}{63.16}
& 50.37 & 55.38 & 43.35
& 53.58 & 53.30
& 18.06 & 28.94 \\
& & LAVT~\cite{yang2022lavt} 
&  \textcolor{gray}{74.46} & \textcolor{gray}{76.89} &  \textcolor{gray}{70.94}
 & 60.64  & 66.19  & 51.87
& 63.45  & 63.33 
 & 23.52  & 39.33 \\
& & Ours & \textcolor{gray}{75.03}  & \textcolor{gray}{77.20}  &\textcolor{gray}{70.10}
    & \textbf{61.60}  & \textbf{67.09}  & \textbf{54.06}
   & \textbf{64.32}  & \textbf{64.70}
   & \textbf{25.75}  & \textbf{40.60} \\
\hline
\multirow{6}*{RefCOCO+}
& \multirow{3}*{oIoU}
& VLT~\cite{ding2021vision}
& 46.35  & 51.74  & 40.73 
& \textcolor{gray}{50.93} & \textcolor{gray}{54.90} &\textcolor{gray}{45.25}
& 48.93 & 50.13
& 14.66 & 22.40 \\
& & LAVT~\cite{yang2022lavt}
& 61.01 & 66.98  & 55.40
 & \textcolor{gray}{62.14} & \textcolor{gray}{68.38} & \textcolor{gray}{55.1}
&  59.06  & 60.86
 & 17.80  & 29.46 \\
& & Ours &\textbf{66.03}  & \textbf{70.03}  & \textbf{58.91} 
    &  \textcolor{gray}{63.35}  &  \textcolor{gray}{70.16} &  \textcolor{gray}{56.70}
   & \textbf{61.95}  & \textbf{64.55}
   & \textbf{20.46}  & \textbf{32.62} \\
\cline{2-13}
& \multirow{3}*{mIoU}
& VLT~\cite{ding2021vision}
& 51.67  & 56.70  & 45.39 
& \textcolor{gray}{55.42} & \textcolor{gray}{59.37} &\textcolor{gray}{49.44}
   & 52.86  & 53.50
   & 13.87  & 22.18 \\
& & LAVT~\cite{yang2022lavt}
&  64.89 & 70.06 & 59.53
 & \textcolor{gray}{65.81} & \textcolor{gray}{70.97} & \textcolor{gray}{59.23}
&  63.20 & 64.20
 & 19.02  & 32.11 \\
& & Ours & \textbf{68.83}  & \textbf{72.73}  & \textbf{62.71}
    & \textcolor{gray}{66.24} & \textcolor{gray}{72.53} & \textcolor{gray}{60.75}
   & \textbf{66.19}  & \textbf{66.90}
   & \textbf{22.48}  & \textbf{36.16} \\
\hline
\multirow{4}*{ReferIt}
& \multirow{2}*{oIoU}
& LAVT~\cite{yang2022lavt}
& 52.93 & 57.01 &  48.68
 & 37.20 & 41.13  & 32.35
&  39.94  & 38.17
 &  \textcolor{gray}{74.60}  &  \textcolor{gray}{71.94} \\
& &Ours&\textbf{54.03}  & \textbf{58.41}  & \textbf{49.92} 
    & \textbf{39.55}  & \textbf{44.20}  & \textbf{34.03}
   & \textbf{42.36}  & \textbf{42.14}
   & \textcolor{gray}{75.61} &\textcolor{gray}{73.78}\\
\cline{2-13}
& \multirow{2}*{mIoU}
& LAVT~\cite{yang2022lavt}
&  54.19 & 57.83 & 49.80
 & 36.99 & 41.66  & 31.39
&  40.12 & 38.95 
 & \textcolor{gray}{66.48} & \textcolor{gray}{63.01} \\
& & Ours & \textbf{55.84}  & \textbf{59.67}  & \textbf{51.30} 
    & \textbf{40.73}  & \textbf{45.86}  & \textbf{34.13} 
   & \textbf{43.64}  & \textbf{42.79}
   & \textcolor{gray}{68.40}  & \textcolor{gray}{65.27} \\
\hline
\multirow{12}*{RefCOCOg}
& \multirow{5}*{oIoU}
& VLT~\cite{ding2021vision}
& 56.91 & 59.42 & 55.94
& 43.34 & 46.01 & 39.21
& \textcolor{gray}{50.77} & \textcolor{gray}{51.58}
& 12.87 & 23.48 \\
& & CRIS~\cite{wang2022cris} 
& 65.77 & 68.73  & 62.37
 & 53.75 & 57.43  & 46.42
&  \textcolor{gray}{57.79}  & \textcolor{gray}{59.06} 
 & 19.59  & 31.95 \\
& & ETRIS~\cite{xu2023bridging}
& 65.29 & 68.60  & 60.28
 & 53.43 & 59.38  & 45.07
&  \textcolor{gray}{59.45}  & \textcolor{gray}{59.66}
 & 16.40  & 24.40 \\
& & CGFormer~\cite{tang2023contrastive}
& 65.73 & 68.73  & 64.80
 & \textbf{57.73} & \textbf{61.38}  & \textbf{51.76}
&  \textcolor{gray}{63.60}  & \textcolor{gray}{64.49}
 & 15.89  & 28.60 \\
\cdashline{3-13}[1pt/1pt]
& & LAVT~\cite{yang2022lavt}
& 65.20 & 67.43  & 63.07
 & 53.88 & 56.52  & 47.45
&  \textcolor{gray}{62.09}  & \textcolor{gray}{60.50} 
 & 17.89  & 29.09 \\

& & Ours & \textbf{69.18}  & \textbf{70.96}  & \textbf{66.65} 
    & 57.64  & 60.74 & 51.13 
   & \textcolor{gray}{63.52}  & \textcolor{gray}{64.00}
   & \textbf{19.88}  & \textbf{32.89} \\
\cline{2-13}
& \multirow{5}*{mIoU}
& VLT~\cite{ding2021vision}
& 59.66 & 61.37 & 59.06
& 46.85 & 50.18 & 43.99
&\textcolor{gray}{54.45} & \textcolor{gray}{54.61}
& 14.65 & 25.58 \\
& & CRIS~\cite{wang2022cris} 
& 68.42 & 70.41  & 65.82
 & 58.46 & 62.00  & 53.26
&  \textcolor{gray}{61.30}  & \textcolor{gray}{61.52} 
 & 21.39  & 33.92 \\
& & ETRIS~\cite{xu2023bridging}
& 65.63 & 68.32  & 62.20
 & 55.93 & 61.39  & 49.39
&  \textcolor{gray}{60.28}  & \textcolor{gray}{60.46}
 & 16.63  & 25.82 \\
& & CGFormer~\cite{tang2023contrastive}
& 67.57 & 70.09  & 67.03
 & 60.56 & 63.98  & \textbf{56.20}
&  \textcolor{gray}{65.63}  & \textcolor{gray}{66.04} 
 & 19.44  & 33.57 \\
\cdashline{3-13}[1pt/1pt]
& & LAVT~\cite{yang2022lavt}
&  66.20 & 67.73 & 64.85
 & 55.88 & 58.25  & 50.47
& \textcolor{gray}{63.62}  & \textcolor{gray}{63.66} 
 & 18.97  & 31.97 \\
& & Ours & \textbf{70.61}  & \textbf{72.13} &\textbf{68.48}
    &\textbf{60.95}  & \textbf{64.00}  &55.99
&\textcolor{gray}{66.49}  & \textcolor{gray}{66.45}
   & \textbf{22.72} & \textbf{37.49} \\
\hline
\end{tabular}
\vspace{-0.3cm}
\label{tab1}
\end{table*}

\noindent\textbf{Evaluation Protocol.} We introduce a zero-shot cross-dataset evaluation protocol to assess the generalization of RIS. The setting among RefCOCO series datasets can evaluate the generalization capabilities to varied textual descriptions since they share image domain and object categories.
To further evaluate the generalization to unseen visual entities, we train on one of the RefCOCO series datasets and test on ReferIt, and vice versa. Notably, due to the small size of the generated GraspNet-RIS, it is exclusively utilized for testing. 

We adopt the common metrics of overall intersection-over-union (oIoU), mean intersection-over-union (mIoU) and precision at 0.5, 0.7, and 0.9 threshold values in the experiments. 
\subsection{Implementation Details}
We maintain the same visual encoder and text encoder as in~\cite{yang2022lavt}. And we follow its training and testing settings without any additional data augmentations or tricks. We adopt ``It is a [] '' as the unified form of the prompt. The MFA is performed on the last PWAM module in LAVT. The aggregation block uses a window size of 5 and the same number of heads as the visual features. The number of output channels in Equation \eqref{aggre} is identical to the input channels, whereas in Equation~\eqref{decoder}, it is one-eighth of the input channels.  We extract vision features from a Swin-Transformer pretrained on ImageNet22k, which is frozen during training\footnote{Using the models with better pixel coherence and stronger generalization can bring more improvements. We adopt it for a fair comparison since it is identical to the initial visual encoder.}.
\begin{figure*}[htb]
\setlength{\abovecaptionskip}{0pt}
\setlength{\belowcaptionskip}{-0pt}
\vspace{5pt}
\centerline{\includegraphics[width=0.91\textwidth]{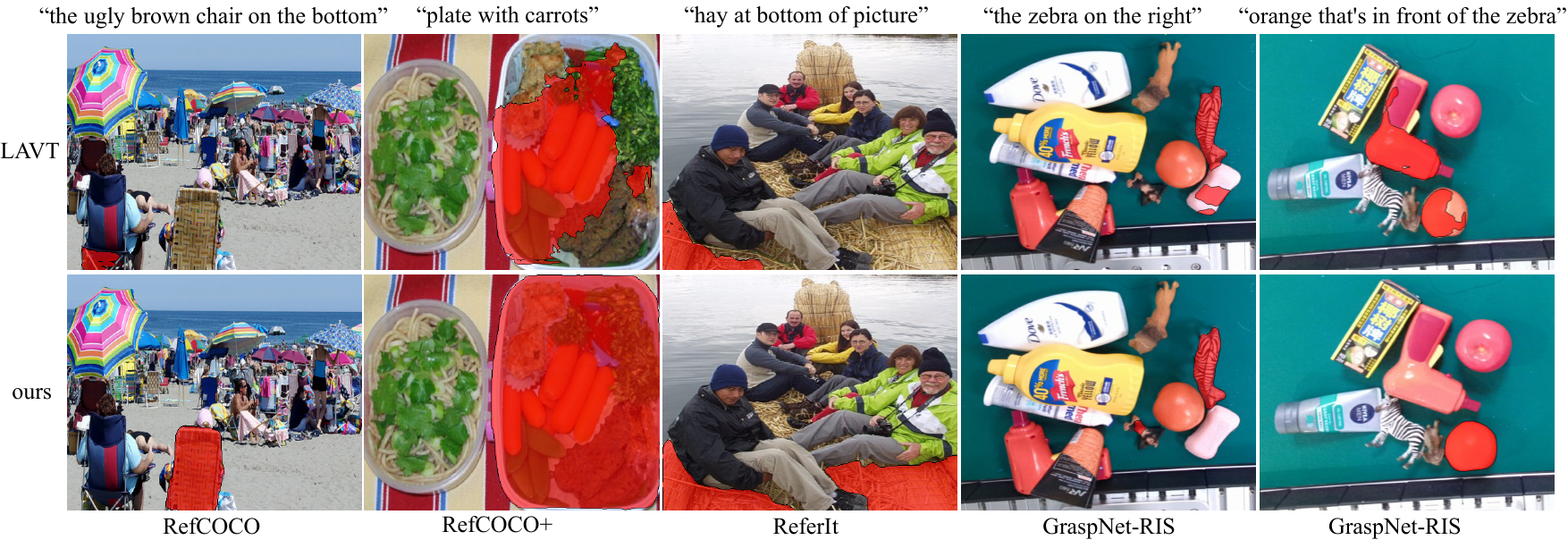}}
\caption{Visualization of predicted masks on examples of the zero-shot generalization from the model with one suit of parameters trained on RefCOCOg.}
\vspace{-0.3cm} 
\label{fig4}
\end{figure*}
\subsection{Comparison in Zero-Shot Cross-Dataset Setting}
We compare our method with the current best-performing methods VLT~\cite{ding2021vision}, LAVT~\cite{yang2022lavt}, ETRIS~\cite{xu2023bridging}, CRIS~\cite{wang2022cris} and CGformer~\cite{tang2023contrastive}. We use the pretrained model released by VLT and LAVT to report the results under the evaluation protocol.
In addition, we train LAVT on ReferIt using the official code and compare it with our method in all settings since we share the same model and training configurations.
We train CRIS, ETRIS, and CGFromer on RefCOCOg using their official code and evaluate the zero-shot performance on other datasets. Our visual encoder and text encoder are identical to those in LAVT and CGformer, and ETRIS adopts more stronger generalizable encoders, CLIP-ViT-B16. Notably, we exclude the multi-scale loss in CGformer and adopt the same loss as LAVT and ours for a fair comparison.  All the models are trained for 50 epochs with batch size 32.

Tabel~\ref{tab1} illustrates the comparison results on the public benchmarks.
Our method consistently improves LAVT across all settings. We concretely analyze the comparison under the setting that the methods are all trained on RefCOCOg and evaluated on other datasets in the following. Compared to LAVT, our method improves the average zero-shot mIoU by 4.15\%, 5.45\%, and 4.64\% on RefCOCO, RefCOCO+ and ReferIt respectively. 
CRIS, a clip-based contrastive learning method, exhibits strong generalization despite the superior supervised performance. Compared to it, our method achieves average improvements of 2.19\%, 2.41\%, 2.45\% mIoU on RefCOCO, RefCOCO+ and ReferIt respectively. ETRIS freezes the CLIP encoders during training, introducing a few additional learnable parameters to better transfer the CLIP knowledge into RIS. Our method surpasses it by  5.02\%, 4.74\%, 8.88\% in terms of average mIoU on RefCOCO, RefCOCO+ and ReferIt respectively.
\begin{table}[h]
\caption{Zero-shot performance on GraspNet-RIS.}
\label{tab2}
\resizebox{\linewidth}{!}{
\begin{tabular}{c|c|c c c c c}
\hline
 \multirow{2}*{Train Dataset} & \multirow{2}*{Method} & \multicolumn{5}{c}{GraspNet-RIS}\\
 \cline{3-7}
 & &  mIoU & oIoU & Prec@0.5 & Prec@0.7 & Prec@0.9\\
\hline
\multirow{2}*{RefCOCO}  & LAVT  & 35.17 & 31.51  & 37.10 & 28.43 & 10.69 \\
& Ours & \textbf{39.27} & \textbf{34.46} & \textbf{41.94} & \textbf{30.65} & \textbf{12.30} \\
\hline
\multirow{2}*{RefCOCO+}  & LAVT  & 37.43 & 32.18  & 39.92 & 29.03 & \textbf{13.51} \\
& Ours & \textbf{38.70} & \textbf{32.41} & \textbf{42.54} & \textbf{30.04} & 11.29 \\
\hline
\multirow{2}*{RefCOCOg}  & LAVT  & 36.90 & 35.10  & 39.72 & 33.87 & 14.31 \\
& Ours & \textbf{44.53} & \textbf{37.06} & \textbf{48.19} & \textbf{37.30} & \textbf{15.52} \\
\hline
\multirow{2}*{ReferIt}  & LAVT  & 38.47 & 33.10  & 40.12 & 26.41 & 7.06 \\
& Ours & \textbf{43.63} & \textbf{37.36} & \textbf{47.58} & \textbf{35.48} & \textbf{8.06} \\
\hline
\end{tabular}
}
\vspace{-0.3cm} 
\end{table}
And our method outperforms the current best method in generalization, CGformer, by 3.6\% and 4.14\% in terms of average mIoU and oIoU on the most challenging zero-shot evaluation on ReferIt.
We exceed CGformer by an average of 2.18\% mIoU on RefCOCO whose expressions are more varied and unconstrained. The example visualizations of the zero-shot predictions from the model with one suit of parameters trained on RefCOCOg are shown in Fig.~\ref{fig4}.

Tabel~\ref{tab2} shows the zero-shot performance on GraspNet-RIS of models trained on four public datasets. Our method achieves clear improvements in all settings, e.g., our method notably outperforms LAVT by 5.16\% mIoU and 4.26\%  oIoU when trained on ReferIt. It demonstrates that our method also generalizes well to scenarios with large domain shifts in practical applications.

\subsection{Ablation Study}
In this section, we conduct ablation studies to evaluate the effects of core components of the proposed model
under the setting that the model is trained on RefCOCOg and evaluated on other datasets. 

\begin{table}[h]
\caption{Ablation results of the zero-shot performance on RefCOCO, RefCOCO+ and ReferIt. The model is trained on RefCOCOg. TP: Target Prompt. MFA: Multi-modal fusion aggregation. VG: Visual guidance.}
\resizebox{\linewidth}{!}{
\begin{tabular}{c|c|c|c c c c c}
\hline
 \multicolumn{2}{c|}{Test Dataset} & Method & mIoU & oIoU & Prec@0.5 & Prec@0.7 & Prec@0.9\\
\hline
\multirow{9}*{RefCOCO} &\multirow{3}*{val} & baseline  & 66.20 & 65.20  & 74.51 & 64.39 & 25.97 \\
  &  & + TP & 68.98 & 67.26 & 76.89 & 68.41 & 34.19 \\
&  &  + MFA\&VG  & \textbf{70.61} & \textbf{69.18} & \textbf{78.52} & \textbf{70.68} & \textbf{35.21} \\
\cline{2-8}
&\multirow{3}*{testA} & baseline  & 67.73 & 67.43  & 77.96 & 67.19 & 24.25 \\
&&+ TP&71.27&69.98&80.56&72.55&35.04\\
&&+ MFA\&VG&\textbf{72.13}&\textbf{70.96}&\textbf{81.24}&\textbf{73.75}&\textbf{35.99}\\
\cline{2-8}
&\multirow{3}*{testB} & baseline  & 64.85 & 63.07  & 71.82 & 59.98 & 27.42 \\
&&+TP&67.10&65.06&73.70&63.61&34.86\\
&&+MFA\&VG&\textbf{68.48}&\textbf{66.65}&\textbf{75.31}&\textbf{65.69}&\textbf{37.57}\\
\hline
\multirow{9}*{RefCOCO+} &\multirow{3}*{val} & baseline  & 55.88 & 53.88  & 61.99 & 52.59 & 20.19 \\
  &  & + TP & 59.38 & 56.33 & 65.63 & 57.70 & 28.68 \\
&  &  + MFA\&VG  & \textbf{60.95} & \textbf{57.64} & \textbf{67.25} & \textbf{59.52} & \textbf{30.16} \\
\cline{2-8}
&\multirow{3}*{testA} & baseline  & 58.25 & 56.52  & 65.68 & 56.23 & 20.31 \\
&&+ TP&63.43&59.77&70.82&63.22&30.42\\
&&+ MFA\&VG&\textbf{64.00}&\textbf{60.74}&\textbf{71.38}&\textbf{64.01}&\textbf{30.89}\\
\cline{2-8}
&\multirow{3}*{testB} & baseline  & 50.47 & 47.45  & 54.12 & 44.16 & 20.37 \\
&&+ TP&54.48&50.30&58.54&50.42&27.59\\
&&+ MFA\&VG&\textbf{55.99}&\textbf{51.13}&\textbf{60.26}&\textbf{52.08}&\textbf{29.56}\\
\hline
\multirow{6}*{ReferIt} &\multirow{3}*{val} & baseline  & 18.97 & 17.89  & 19.09 & 13.30 & 3.29 \\
  &  & + TP & 19.68 & 16.29 & 19.73 & 13.73 & 2.98 \\
&  &  + MFA\&VG  & \textbf{22.72} & \textbf{19.88} & \textbf{22.70} & \textbf{15.62} & \textbf{3.49} \\
\cline{2-8}
&\multirow{3}*{test} & baseline  & 31.97 & 29.09  & 34.43 & 25.27 & 4.95 \\
&&+ TP&33.97&29.09&37.39&26.68&4.97\\
&&+ MFA\&VG&\textbf{37.49}&\textbf{32.89}&\textbf{41.04}&\textbf{29.93}&\textbf{5.56}\\

\hline
\end{tabular}
}
\vspace{-0.4cm}
\label{tab3}
\end{table}
In Tabel~\ref{tab3}, the first row of each section shows the baseline for comparison, which is the results of the model released by LAVT. Boosting the referring expression with the target prompt achieves a significant improvement on RefCOCO series datasets, with an average increase of 2.85\% mIoU,  8.82\%  Prec@0.9 on RefCOCO, and 4.23\% mIoU, 8.61\% Prec@0.9 on RefCOCO+, which demonstrates that the target prompt significantly enhances the model's ability to generalize to various unconstrained textual descriptions.

Introducing MFA and visual guidance further improves the zero-shot performance. 
The model attains noticeable improvements over the baseline with wide average margins of 4.15\% mIoU, 10.38\%  Prec@0.9 on RefCOCO, and 
of 5.45\% mIoU, and 9.91\%  Prec@0.9 on RefCOCO+. The zero-shot performance on ReferIt is relatively poor since it contains numerous unseen visual entities.
As expected, MFA and visual guidance provide more benefits to ReferIt, while target prompt yields greater improvements on RefCOCO and RefCOCO+. The model with MFA and visual guidance achieves improvements of 3.75\% mIoU and 5.52\% mIoU on the validation and test subsets of ReferIt respectively.

\section{CONCLUSION}
In this paper, we propose to improve the generalization of RIS from two perspectives. To robustly generalize to various unconstrained textual descriptions, we boost a given expression with an explicit target prompt in a unified context to facilitate the target capturing. To deal with the incomplete targets and false-positive clumps of predictions in unseen visual entities, we introduce a multi-modal fusion aggregation module along with visual guidance from a powerful pretrained model to leverage the spatial relations and visual coherences. Extensive experiments conducted under the zero-shot cross-dataset evaluation protocol demonstrate the effectiveness of the proposed approach.

\bibliography{reference.bib} 
\bibliographystyle{IEEEtran} 

\end{document}